\documentclass{article}
\usepackage{spconf,amsmath,amssymb,booktabs,graphicx,multirow}
\usepackage{url}
\usepackage{xcolor}
\usepackage{algorithm,algorithmic}
\usepackage{etoolbox}

\providecommand{\LONGVERSION}{1}
\newif\iflongversion
\ifnum\LONGVERSION=1\relax
  \longversiontrue
\else
  \longversionfalse
\fi

\newcommand{\versiontext}[2]{\iflongversion#1\else#2\fi}

\makeatletter
\patchcmd{\@makecaption}{\vskip 10pt}{\vskip 4pt}{}%
  {\PackageError{paper-spacing}{Could not patch spconf caption spacing}{Check spconf.sty.}}
\makeatother

\makeatletter
\renewcommand\section{\@startsection{section}{1}{\z@}%
  {-10pt}{5pt}{\normalfont\Large\bfseries}}
\renewcommand\subsection{\@startsection{subsection}{2}{\z@}%
  {-8pt}{4pt}{\normalfont\large\bfseries}}
\makeatother

\title{Privacy-Aligned Personalized Federated Learning with Compact Adaptation and Variable-Length Gaussian Communication}

\name{Yilin Xu$^{1}$, Chun Hei Michael Shiu$^{2}$, Chih Wei Ling$^{3}$, Linqi Song$^{1}$}
\address{
  $^{1}$Department of Computer Science, City University of Hong Kong \\
  $^{2}$Department of Electrical and Computer Engineering, University of British Columbia \\
  $^{3}$School of Computer Science and Engineering, Hebrew University of Jerusalem
}

\begin{document}
\ninept
\flushbottom
\maketitle

\setlength{\abovedisplayskip}{5pt}
\setlength{\belowdisplayskip}{5pt}
\setlength{\abovedisplayshortskip}{0pt}
\setlength{\belowdisplayshortskip}{4pt}

\begin{abstract}
Record-level differential privacy exposes a structural misalignment in personalized federated learning when client-specific variation is low-dimensional while training repeatedly releases high-dimensional updates. 
In this paper, we address this misalignment by releasing a private client context once and confining repeated adaptation to a fixed coefficient space. Beyond dimensionality reduction, the factorized generator induces an adaptive optimization geometry that reshapes noisy updates, and controlled ablations show that most of its private-training gain is retained by radial evolution. 
To further reduce the communication cost, we realize the Gaussian mechanism for coefficient updates directly through variable-length quantization with finite expected code length,
so that the quantization error itself serves as the required privacy perturbation rather than extra distortion.
Across MNIST and CIFAR-10, our design matches or outperforms full-model private adaptation across privacy budgets and client heterogeneity, while reducing protected uplink by a factor of 2.67 at \(\varepsilon=16\) on CIFAR-10 with comparable future-client accuracy.
\end{abstract}

\begin{keywords}
federated learning, differential privacy, randomized quantization
\end{keywords}

\section{Introduction}
\label{sec:intro}

Personalized federated learning (PFL) addresses statistical heterogeneity without restricting all clients to a single global model. Hypernetwork-based PFL maps a compact client context to personalized model parameters \cite{pfedhn,odpfl,pefll,hypflzero}. When a suitable context is available, this formulation allows previously unseen clients to obtain personalized models without iterative local fine-tuning, making this formulation attractive for future-client personalization.

However, differential privacy (DP) changes this picture. Several recent works have explored private personalization with meta-learning, hypernetworks, and shared representations \cite{wei2023dppfl,nemala2023dphyper,snedeker2025private}. Nevertheless, future-client personalization remains challenging because trainable client embeddings require adaptation for previously unseen clients \cite{pfedhn,nemala2023dphyper}, while learned context encoders \cite{pefll,hypflzero} add another data-dependent optimization path that must be privatized. A direct privatization of hypernetwork PFL may repeatedly release gradients of the generated high-dimensional model. Although the privacy parameter of an $\ell_2$-bounded Gaussian mechanism is not set by dimension alone, both the expected perturbation energy and the communication overhead grow with the dimension of the released signal. Low-dimensional private optimization and reparameterization have been used to reduce this ambient-dimensional burden \cite{zhou2021projected,yu2021rgp}, but in hypernetwork PFL this issue is coupled to client representation and future-client adaptation.

This motivates a simple question. Can client-specific information be released only once while repeated private adaptation is confined to a compact space? We answer this by replacing iterative client representation learning with a one-shot private context and moving repeated private adaptation from the full model to a fixed coefficient space. In addition, a factorized server-side generator maps the cached context to personalized coefficients and changes how noisy coefficient updates propagate through shared parameters.
We further reduce the communication cost of repeated private updates with layered rejection-sampled universal quantization (LRSUQ) \cite{ling2025lrsuq}. Under our trusted-server setting, 
LRSUQ realizes the prescribed decoded Gaussian channel using variable-length messages with finite expected code length, and the privacy perturbation itself serves as quantization error.
Together, the coefficient-space parameterization and variable-length Gaussian channel retain one-shot personalization for unseen clients while reducing the dimensional and communication burden of private adaptation.

\textbf{Contributions.}
(i) We reformulate record-private hypernetwork PFL around a one-shot private client context and coefficient-space adaptation, enabling personalization of unseen clients without iterative local fine-tuning. (ii) We characterize the optimization geometry induced by the factorized generator and show with controlled ablations that radial evolution retains most of the factorization gain under private training. (iii) We instantiate coefficient updates with LRSUQ, which by construction realizes the prescribed decoded Gaussian law using variable-length messages with finite expected code length, and show experimentally that it reduces protected uplink with essentially unchanged utility.

\section{Method}
\label{sec:method}
Our method includes three components. We first define one-shot private conditioning and repeated coefficient-space adaptation. Then, we analyze the geometry induced by the factorized generator and finally replace dense Gaussian coefficient messages by a variable-length realization of the same decoded channel.

\subsection{Private Contextual Adaptation}
\label{sec:private-adaptation}

We consider central record-level DP with a trusted server \cite{dwork2014algorithmic}. A record is
$z=(x,y)$, where $x$ is the input and $y$ is its label.  
For the federated dataset $\mathcal D=(D_1,\ldots,D_N)$, \emph{replace-one adjacency} $\mathcal D\sim\mathcal D'$ means $D_i'=(D_i\setminus\{z\})\cup\{z'\}$ for one client $i$ and $D_j'=D_j$ for $j\neq i$.
To capture client-specific information without learning a separate private encoder, we construct a bounded context from first- and second-order input statistics. Let $\{\mathcal G_j\}_{j=1}^{s}$ be a fixed partition of the input coordinates into $s$ feature groups, and define
\begin{align}
    \mu_j(x)
    &=
    \frac{1}{|\mathcal G_j|}
    \sum_{u\in\mathcal G_j} x_u,
    \quad
    \nu_j(x)
    =
    \frac{1}{|\mathcal G_j|}
    \sum_{u\in\mathcal G_j} x_u^2,
    \nonumber\\
    \phi(x)
    &=
    \frac{1}{\sqrt{2s}}
    \big[
        \tanh(\mu(x))^\top,\,
        \tanh(\nu(x))^\top
    \big]^\top .
    \label{eq:context-map}
\end{align}
Here $\mu(x),\nu(x)\in\mathbb R^s$, so $\phi(x)\in\mathbb R^{2s}$ and $\|\phi(x)\|_2\leq1$.
For the image experiments, the groups $\mathcal G_j$ correspond to the input
channels.
Let $\mathcal C_i\subset D_i$ be a fixed \emph{conditioning set} of $m_c$ records used to form the one-shot context statistic.
The client releases the empirical context information only once through
\begin{equation}
    \widetilde c_i
    =
    \frac{1}{m_c}
    \sum_{(x,y)\in\mathcal C_i}
    \phi(x)
    +
    \zeta_i,
    \qquad
    \zeta_i\sim
    \mathcal N(0,\sigma_c^2 I).
    \label{eq:private-context}
\end{equation}
Because $\|\phi(x)\|_2\leq1$, the one-shot context release has replace-one sensitivity at most $2/m_c$. The server then caches $\widetilde c_i$, so an unseen client requires only this one-shot private context release for subsequent model generation.
Conditioned on $\widetilde c_i$, the server generates a personalized model through the factorized coefficient map
\begin{equation}
    h_i=h_\psi(\widetilde c_i),
    \qquad
    a_i=Wh_i,
    \qquad
    \theta_i=\theta_{\rm ref}+Pa_i .
    \label{eq:model-generator}
\end{equation}
Here $h_\psi:\mathbb R^{2s}\rightarrow\mathbb R^r$ is a compact multilayer perceptron (MLP), $W\in\mathbb R^{k\times r}$ is a trainable matrix mapping its hidden representation $h_i$ to $k$-dimensional coefficient space, and $\theta_i\in\mathbb R^{d_\theta}$ is the resulting personalized model. The reference parameters $\theta_{\rm ref}\in\mathbb R^{d_\theta}$ and coefficient basis $P\in\mathbb R^{d_\theta\times k}$ are initialized once using public randomness and then kept fixed. The columns of $P$ form an orthonormal basis of the $k$-dimensional adaptation subspace $\mathcal U=\operatorname{range}(P)\subset\mathbb R^{d_\theta}$, with $P^\top P=I_k$. \versiontext{%
Thus, $P$ fixes a data-independent adaptation space, while $(W,\psi)$ learn context-dependent models within it (see \mbox{Appendix~\ref{app:coefficient-space}}).%
}{%
Thus, $P$ defines a reproducible adaptation space independently of client data, while $W$ and $\psi$ learn how each private context determines a personalized model within that space.%
}

During training, selected clients differentiate their local loss with respect to the coefficients rather than the full model parameters. For a record $z$, define
\begin{equation}
    g_i(z)
    =
    \left.
    \nabla_a
    \ell\!\left(\theta_{\rm ref}+Pa;\,z\right)
    \right|_{a=a_i}.
    \label{eq:record-gradient}
\end{equation}
At communication round $t$, let $\mathcal B_{i,t}\subset D_i$ contain $m_g$ records and
define $\operatorname{clip}_{C_g}(v)=v\min\{1,C_g/\|v\|_2\}$. Client $i$ releases
\begin{equation}
    \widetilde g_{i,t}
    =
    \frac{1}{m_g}
    \sum_{z\in\mathcal B_{i,t}}
    \operatorname{clip}_{C_g}\!\left(g_i(z)\right)
    +
    \xi_{i,t},
    \quad
    \xi_{i,t}\sim\mathcal N(0,\sigma_g^2I_k).
    \label{eq:private-gradient}
\end{equation}
The fixed denominator and clipping imply replace-one sensitivity at most $2C_g/m_g$.
Let $\omega=(W,\psi)$ denote the trainable generator parameters, and write $J_i=\partial a_i/\partial\omega$ for the Jacobian of the generated coefficients with respect to the trainable generator parameters. Let $\mathcal S_t$ denote the participating clients selected for the server update at round $t$.
With uniform client weighting, the server update is
\begin{equation}
    \omega_{t+1}
    =
    \omega_t
    -
    \frac{\eta}{|\mathcal S_t|}
    \sum_{i\in\mathcal S_t}
    J_i^\top \widetilde g_{i,t}.
    \label{eq:server-update-marked}
\end{equation}

\begin{algorithm}[t]
\caption{Privacy-aligned personalized federated learning}
\label{alg:overview}
\begin{algorithmic}[1]
\STATE Publicly initialize $\theta_{\rm ref}$ and the matrix $P$.
\STATE For each participating client $i$, release $\widetilde c_i$ once using Eq.~\eqref{eq:private-context} and cache it at the server.
\FOR{round $t=1,\ldots,T$}
    \STATE Select the participating clients $\mathcal S_t$ for round $t$.
    \FOR{$i\in\mathcal S_t$}
        \STATE Generate $a_i=Wh_\psi(\widetilde c_i)$ and $\theta_i=\theta_{\rm ref}+Pa_i$.

        \STATE Compute coefficient gradients $g_i(z)$ using Eq.~\eqref{eq:record-gradient}.

        \STATE Clip and average the per-record gradients and realize the Gaussian release $\widetilde g_{i,t}$ in Eq.~\eqref{eq:private-gradient}.
    \ENDFOR
    \STATE Update $(W,\psi)$ using Eq.~\eqref{eq:server-update-marked}.
\ENDFOR
\STATE For an unseen client, release its context once and generate its model with no local fine-tuning.
\end{algorithmic}
\end{algorithm}

Algorithm~\ref{alg:overview} summarizes the complete training and deployment procedure.
Privacy is accounted for using R\'enyi differential privacy (RDP) \cite{Mironov17}. Let $\Delta_c=2/m_c$ and $\Delta_g=2C_g/m_g$ denote the sensitivities of the one-shot context and repeated coefficient releases.
We use a common Gaussian standard deviation $\sigma_c=\sigma_g=\sigma$. For a client participating in $T_i$ private update rounds, composition of the one-shot context release and its repeated coefficient-gradient releases gives
\begin{equation}
A_i=\frac{\Delta_c^2+T_i\Delta_g^2}{2\sigma^2},
\qquad
\varepsilon_i=A_i+2\sqrt{A_i\log(1/\delta)} .
\label{eq:privacy-accounting}
\end{equation}
\versiontext{%
Optimizing the RDP-to-$(\varepsilon,\delta)$ conversion over continuous R\'enyi orders gives the second expression; we report $\varepsilon=\max_i\varepsilon_i$ (see \mbox{Appendix~\ref{app:privacy}}).%
}{%
The second expression is obtained by optimizing the standard RDP-to-$(\varepsilon,\delta)$ conversion over continuous Rényi orders, and we report $\varepsilon=\max_i\varepsilon_i$.%
}

\begin{table*}[!t]
\centering
\caption{Accuracy (\%) across privacy budgets. The upper and lower blocks report accuracy on participating clients (Seen) and query accuracy on unseen clients (Future), respectively. Bold denotes the largest private mean in each column.}
\label{tab:privacy_results}
\setlength{\tabcolsep}{2.05pt}
\begin{tabular}{@{}lcccccccc@{}}
\toprule
& \multicolumn{4}{c}{MNIST} & \multicolumn{4}{c}{CIFAR-10} \\
\cmidrule(lr){2-5}\cmidrule(lr){6-9}
Method & $\varepsilon=4$ & $\varepsilon=8$ & $\varepsilon=16$ & $\varepsilon=32$ & $\varepsilon=4$ & $\varepsilon=8$ & $\varepsilon=16$ & $\varepsilon=32$ \\
\midrule
\multicolumn{9}{l}{\textit{Seen clients}} \\
Global &
$22.98\!\pm\!8.62$ & $31.30\!\pm\!9.30$ & $44.10\!\pm\!5.83$ & $55.63\!\pm\!2.85$ &
$12.29\!\pm\!0.81$ & $15.06\!\pm\!2.94$ & $17.94\!\pm\!2.54$ & $21.47\!\pm\!3.08$ \\
Full-DG &
$66.06\!\pm\!3.86$ & $78.59\!\pm\!2.44$ & $86.42\!\pm\!0.99$ & $90.66\!\pm\!2.31$ &
$37.29\!\pm\!1.98$ & $43.42\!\pm\!1.64$ & $47.98\!\pm\!1.81$ & $50.46\!\pm\!1.23$ \\
DeepLin-DG &
$65.35\!\pm\!3.93$ & $79.37\!\pm\!1.23$ & $85.78\!\pm\!1.23$ & $89.34\!\pm\!1.59$ &
$38.50\!\pm\!1.83$ & $44.45\!\pm\!1.69$ & $49.18\!\pm\!0.83$ & $51.49\!\pm\!1.11$ \\
Ours-DG &
$68.76\!\pm\!4.56$ & $\mathbf{81.28\!\pm\!3.17}$ & $\mathbf{88.39\!\pm\!2.62}$ & $91.64\!\pm\!2.07$ &
$\mathbf{38.83\!\pm\!2.02}$ & $\mathbf{45.11\!\pm\!2.66}$ & $\mathbf{50.66\!\pm\!0.97}$ & $53.81\!\pm\!0.99$ \\
Ours-LRSUQ &
$\mathbf{69.55\!\pm\!4.43}$ & $80.39\!\pm\!4.13$ & $86.69\!\pm\!2.83$ & $\mathbf{91.82\!\pm\!0.82}$ &
$38.77\!\pm\!1.63$ & $45.05\!\pm\!3.29$ & $50.33\!\pm\!2.37$ & $\mathbf{54.31\!\pm\!0.97}$ \\
\midrule
\multicolumn{9}{l}{\textit{Future clients}} \\
Global &
$21.67\!\pm\!13.50$ & $24.81\!\pm\!10.17$ & $35.09\!\pm\!5.43$ & $47.40\!\pm\!4.99$ &
$8.42\!\pm\!5.21$ & $9.46\!\pm\!4.86$ & $12.77\!\pm\!4.21$ & $15.88\!\pm\!3.86$ \\
Full-DG &
$65.21\!\pm\!2.68$ & $76.85\!\pm\!2.09$ & $85.13\!\pm\!1.61$ & $89.54\!\pm\!2.77$ &
$34.44\!\pm\!3.33$ & $40.99\!\pm\!3.27$ & $45.81\!\pm\!4.05$ & $48.25\!\pm\!4.11$ \\
DeepLin-DG &
$63.44\!\pm\!7.40$ & $78.43\!\pm\!1.31$ & $84.76\!\pm\!1.06$ & $88.97\!\pm\!1.55$ &
$35.62\!\pm\!1.72$ & $40.25\!\pm\!0.90$ & $45.70\!\pm\!2.33$ & $48.42\!\pm\!2.67$ \\
Ours-DG &
$\mathbf{65.23\!\pm\!6.95}$ & $\mathbf{80.09\!\pm\!3.86}$ & $\mathbf{87.56\!\pm\!2.05}$ & $90.99\!\pm\!2.00$ &
$36.69\!\pm\!1.29$ & $43.14\!\pm\!2.78$ & $48.84\!\pm\!3.86$ & $52.33\!\pm\!3.56$ \\
Ours-LRSUQ &
$64.20\!\pm\!5.14$ & $78.15\!\pm\!5.37$ & $85.57\!\pm\!3.25$ & $\mathbf{91.36\!\pm\!1.20}$ &
$\mathbf{36.99\!\pm\!1.08}$ & $\mathbf{43.67\!\pm\!3.08}$ & $\mathbf{48.97\!\pm\!3.18}$ & $\mathbf{52.79\!\pm\!3.25}$ \\
\bottomrule
\end{tabular}
\end{table*}

\begin{figure*}[!t]
\centering
\includegraphics[width=0.94\textwidth]{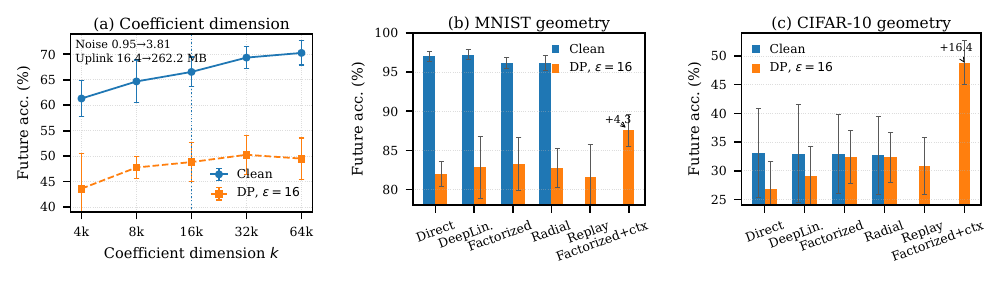}
\caption{Mechanism-focused ablations. (a) Coefficient-dimension trade-off on
CIFAR-10. (b)--(c) Geometry and conditioning controls on MNIST and CIFAR-10.}
\label{fig:triptych}
\end{figure*}

\subsection{Factorized Optimization Geometry}
\label{sec:geometry}

From Eq.~\eqref{eq:server-update-marked}, client $i$'s contribution to the server update is $\Delta\omega^{(i)}=-(\eta/|\mathcal S_t|)J_i^\top\widetilde g_{i,t}$. Linearizing the coefficients generated for client $j$ gives
\begin{equation}
\Delta a_j^{(i)}
\approx
-\frac{\eta}{|\mathcal S_t|} K_{ji}\widetilde g_{i,t},
\qquad
K_{ji}=J_jJ_i^\top .
\label{eq:induced-kernel}
\end{equation}
Hence, $K_{ji}$ is the coefficient-space kernel induced by the generator and determines how a noisy private update from client $i$ is transferred through shared parameters to client $j$.  
Since $a_i=Wh_i$ and $\omega=(W,\psi)$, direct differentiation with respect to the two parameter blocks yields
\begin{equation}
    K_{ji}
    =
    \langle h_j,h_i\rangle I_k
    +
    W H_jH_i^\top W^\top .
    \label{eq:kernel-decomposition}
\end{equation}
Here $H_i=\frac{\partial h_i}{\partial\psi}$. The first term arises from updating $W$ and acts as an isotropic gain in coefficient space, whereas the second comes from updating MLP parameters $\psi$ and introduces a direction-dependent correction of rank at most $r$. For $i=j$, the first term becomes $\|h_i\|_2^2I_k$, showing that the hidden-state norm
acts as an adaptive scalar gain on the coefficient update.

Writing $h_i=\rho_i v_i$ with $\|v_i\|_2=1$ makes this interpretation more
explicit. If the hidden direction $v_i$ remains fixed while only its magnitude
$\rho_i$ changes, the isotropic term varies as $\rho_i^2 I_k$ without
introducing directional motion in the hidden representation. We refer to this
behavior as \emph{radial evolution}. \versiontext{%
This motivates the Radial and Scalar Replay controls in Sec.~\ref{sec:geometry-ablation}, which distinguish magnitude adaptation from hidden-direction evolution; \mbox{Appendix~\ref{app:geometry}} derives the kernel and explains these controls.%
}{%
This observation motivates the Radial and
Scalar Replay controls in Sec.~\ref{sec:geometry-ablation}, which test whether
the factorization gain is explained primarily by hidden-state magnitude
adaptation or requires the full direction-dependent geometry.%
}

\subsection{Variable-Length Gaussian Communication via LRSUQ}
\label{sec:lrsuq}
Because the client context is communicated only once, repeated coefficient gradients dominate the uplink. We therefore seek a variable-length representation that preserves the prescribed Gaussian mechanism after decoding while having finite expected code length. Let $L$ denote the length, in bits, of the encoded message, and the communication requirement is $\mathbb E[L]<\infty$, where the expectation is over the randomness of the encoding procedure. For a vector-valued statistic $u\in\mathbb R^{d_u}$ with prescribed sensitivity, the direct Gaussian mechanism releases $\mathcal M_{\rm G}(u)=u+\xi$ with $\xi\sim\mathcal N(0,\sigma^2 I_{d_u}).$
Sending this realization directly requires a dense $d_u$-dimensional floating-point vector, which we refer to as \emph{Direct Gaussian (DG)}. Related exact-error and dithered compressors have shown how quantization can be matched to Gaussian privacy noise in federated learning \cite{hegazy2024exact,hasircioglu2024dithering}. LRSUQ additionally admits a native vector-quantization formulation of additive-channel simulation \cite{ling2025lrsuq}, allowing coefficient blocks to be encoded jointly while preserving the prescribed input-independent Gaussian error distribution. This structure is naturally suited to our coefficient-space updates. We therefore apply LRSUQ blockwise, using block size $b=4$ throughout, to instantiate the Gaussian mechanism. The resulting variable-length message decodes to
\begin{equation}
\widehat u
=
\operatorname{Dec}_{\sigma}
\!\left(
\operatorname{Enc}_{\sigma}(u;\mathcal R);
\mathcal R
\right)
=
u+e,
\quad
e\sim\mathcal N(0,\sigma^2 I_{d_u}),
\label{eq:lrsuq-channel}
\end{equation}
where $\mathcal R$ denotes randomness shared by the encoder and decoder, and $e\perp u$. Since $e$ is exactly Gaussian and independent of $u$ \cite{zamir2014lattice,ling2025lrsuq}, the decoded LRSUQ output follows the same additive Gaussian channel, preserving the same privacy accounting and learning dynamics as DG while replacing dense floating-point transmission with a variable-length representation of finite expected code length. Thus, rather than adding separate quantization distortion after privatization, LRSUQ realizes the required Gaussian perturbation through its quantization error. \versiontext{%
With decoding and shared randomness inside the trust boundary, central DP protects decoded releases and model outputs, not raw encoded transcripts (see \mbox{Appendix~\ref{app:lrsuq}}).%
}{%
Under our trusted-server model, decoding and shared randomness remain inside the trust boundary, so the central-DP guarantee applies to decoded releases and subsequent model outputs, not to the raw encoded transcript.%
}

\section{Experiments}
\label{sec:experiments}
The experiments address three questions aligned with our contributions:
whether coefficient-space private adaptation improves the privacy--utility
trade-off, what aspect of the factorized geometry drives the gain, and whether
LRSUQ reduces communication while preserving private-learning utility.

\subsection{Experimental Setup}
Experiments use MNIST \cite{lecun1998gradient} and CIFAR-10 \cite{krizhevsky2009learning}, with examples partitioned into equal-size client datasets using a Dirichlet label partition
\cite{hsu2019nonidentical} with concentration $\alpha=0.5$. Each run uses 40 participating clients for training and 10 previously unseen clients for evaluation. Participating-client records are split $4{:}1$ into private training and evaluation data. 
For every client, the private context is constructed once from $m_c=333$ unlabeled records that are disjoint from those used for accuracy evaluation. Unseen clients are evaluated directly on a disjoint query set without local fine-tuning.

We use a LeNet-style personalized model for both datasets, with \(d_\theta=85{,}822\) on MNIST and \(d_\theta=121{,}182\) on CIFAR-10. The generator hidden dimension is fixed to \(r=100\), and the default coefficient dimension is \(k=16{,}384\), so \(W\in\mathbb R^{16{,}384\times100}\).
We report mean $\pm$ standard deviation over five seeds (41--45) after averaging client accuracy within each seed. 

All methods are trained with stochastic gradient descent (SGD) for 500 communication rounds, using learning rates $0.1$ on MNIST and $0.2$ on CIFAR-10. Each round selects $|\mathcal S_t|=2$ clients. Throughout, $m_c=m_g=333$, $C_g=1$, and $\delta=10^{-5}$. Client schedules are sampled independently of the data and fixed before training. For each seed, the shared noise scale $\sigma$ is calibrated from Eq.~\eqref{eq:privacy-accounting} using the maximum participation count in that fixed schedule, $T_{\max}=\max_iT_i\in[37,47]$ with no subsampling amplification. We evaluate $\varepsilon\in\{4,8,16,32\}$.

\textbf{End-to-end comparisons.}
Five matched variants probe the contribution of each design choice. \emph{Global} is a record-private shared model;
\emph{Full-DG} keeps the same one-shot conditioning but directly generates a displacement in the full $d_\theta$-dimensional model space and privatizes the corresponding full-model gradients, without the fixed coefficient basis $P$; \emph{DeepLin-DG} moves the release to coefficient space with the deep-linear map $a_i=WV\widetilde c_i$; \emph{Ours-DG} uses the full factorized generator $a_i=Wh_\psi(\widetilde c_i)$; and \emph{Ours-LRSUQ} changes only the communication realization of the Gaussian coefficient channel. Accordingly, Full-DG versus Ours-DG evaluates the end-to-end effect of replacing full-model private adaptation with coefficient-space adaptation; DeepLin-DG versus Ours-DG isolates the effect of the nonlinear factorized generator within the same coefficient space; and Ours-DG versus Ours-LRSUQ isolates the communication realization. Because existing private PFL methods often differ in privacy unit,
unseen-client protocol, or adaptation procedure, we focus on matched controls
under a common record-level accountant and future-client evaluation protocol.

\subsection{Privacy-Aligned Personalization}

Table~\ref{tab:privacy_results} reports held-out accuracy on participating clients (\emph{Seen}) and query accuracy on previously unseen clients (\emph{Future}). Across all privacy budgets, coefficient-space personalization matches or outperforms the Full-DG baseline. On Future clients, Ours-DG gains $2.15$--$4.08$ percentage points over Full-DG on CIFAR-10 and up to $3.24$ percentage points on MNIST. Ours-DG also exceeds DeepLin-DG at every evaluated budget, with gains of $1.07$--$3.91$ percentage points on CIFAR-10 and $1.66$--$2.80$ on MNIST.

\textbf{Sensitivity to client heterogeneity.}
At $\varepsilon=16$, Ours-DG consistently outperforms Full-DG on unseen clients across all tested Dirichlet concentrations (Table~\ref{tab:heterogeneity}). The gains range from $2.08$ to $2.98$ percentage points on MNIST and from $1.47$ to $5.41$ points on CIFAR-10, indicating that the coefficient-space advantage is not limited to the default $\alpha=0.5$ partition.

\begin{table}[t]
\centering
\caption{Future client accuracy (\%) at $\varepsilon=16$ across different $\alpha$.}
\label{tab:heterogeneity}
\setlength{\tabcolsep}{2.2pt}
\begin{tabular}{@{}ccccc@{}}
\toprule
& \multicolumn{2}{c}{MNIST}
& \multicolumn{2}{c}{CIFAR-10} \\
\cmidrule(lr){2-3}\cmidrule(lr){4-5}
$\alpha$
& Full-DG & Ours-DG
& Full-DG & Ours-DG \\
\midrule
\multicolumn{5}{l}{\textit{Future clients}} \\
0.8
& $85.88\!\pm\!1.87$ & $\mathbf{88.80\!\pm\!2.25}$
& $43.83\!\pm\!1.28$ & $\mathbf{45.30\!\pm\!1.32}$ \\
0.5
& $85.13\!\pm\!1.61$ & $\mathbf{87.56\!\pm\!2.05}$
& $45.81\!\pm\!4.05$ & $\mathbf{48.84\!\pm\!3.86}$ \\
0.2
& $88.55\!\pm\!2.44$ & $\mathbf{91.53\!\pm\!2.28}$
& $60.81\!\pm\!2.66$ & $\mathbf{63.08\!\pm\!1.86}$ \\
0.1
& $92.60\!\pm\!1.57$ & $\mathbf{94.68\!\pm\!2.27}$
& $70.10\!\pm\!3.84$ & $\mathbf{75.51\!\pm\!2.56}$ \\
\bottomrule
\end{tabular}
\end{table}

\textbf{Coefficient-dimension trade-off.}
At $k=16{,}384$, CIFAR-10 Future accuracy is only $1.43$ percentage points below $k=32{,}768$, while the protected uplink is halved ($65.55$ vs.\ $131.09$ MB) and the measured perturbation norm falls from $2.70$ to $1.91$. Further increasing the dimension to $k=65{,}536$ yields only another $0.68$ percentage points over the default setting but requires four times its uplink. These results place $k=16{,}384$ near a favorable capacity--privacy--communication operating point (Fig.~\ref{fig:triptych}(a)).

\subsection{Factorized Geometry and Client Conditioning}
\label{sec:geometry-ablation}
This experiment separates the optimization effect of factorization from the personalization signal carried by the client context. All geometry controls replace the private context with the same public vector $u_0$. \emph{Direct} optimizes the coefficient vector itself and serves as the unfactorized reference. \emph{DeepLin} uses $a=WVu_0$, retaining a linear factorization without the nonlinear trunk. \emph{Factorized} uses the full map $a=Wh_\psi(u_0)$. \emph{Radial} keeps the initial hidden direction fixed and allows only its magnitude to evolve, testing whether directional motion is necessary. \emph{Replay} applies to Direct the per-round scalar gain $\gamma_t=\|h_t\|_2^2$ recorded from Factorized, testing whether factorization is equivalent to a frozen scalar step-size schedule. \emph{Factorized+ctx} restores the actual one-shot private context and therefore measures the additional contribution of client conditioning.

The controls yield a consistent attribution. Without privacy noise, Direct, DeepLin, Factorized, and Radial reach similar accuracy, so factorization provides little clean-optimization advantage in this setting. At $\varepsilon=16$, Factorized exceeds Direct by $1.19$ percentage points on MNIST and $5.63$ on CIFAR-10. Radial remains within $0.43$ and $0.10$ percentage points of Factorized, respectively, showing that most of the private-training gain is retained without hidden-direction evolution. Replay recovers much of the CIFAR-10 gain but not the MNIST gain, so the effect is not generally explained by a fixed scalar learning-rate schedule alone. Finally, restoring client context adds $4.33$ percentage points on MNIST and $16.39$ on CIFAR-10. This separates two complementary effects. Factorization improves noisy optimization even without client information, while the one-shot context supplies the client-specific signal required for personalization.

\subsection{Variable-Length Gaussian Communication}
This experiment tests whether the exact decoded Gaussian channel can be communicated with substantially fewer bits without changing the private-learning behavior. Because LRSUQ preserves the decoded Gaussian law by construction, Ours-LRSUQ closely tracks Ours-DG across all privacy budgets. On CIFAR-10 at $\varepsilon=16$, Table~\ref{tab:codec_proposed_marked} shows that LRSUQ reduces the coefficient rate from $32.00$ to $12.01$ bpc and the protected uplink from $65.55$ to $24.59$ MB, a $2.67\times$ reduction, while Future accuracy remains essentially unchanged. At approximately $12$ bpc, DG+Q12 and LRSUQ provide comparable utility at nearly identical rates. The distinction is structural. DG+Q12 quantizes an already Gaussian-privatized update, whereas LRSUQ uses quantization error itself to realize the target Gaussian perturbation and therefore preserves the same decoded private-learning channel. 
This communication saving comes with additional encoding–decoding cost, where LRSUQ requires $13.97$ ms per event, while DG+Q12 and DG FP32 require $11.09$ ms and $0.14$ ms per event, respectively.

\begin{table}[!htb]
\centering
\caption{Coefficient communication on CIFAR-10 at
$\varepsilon=16$. ``bpc'' denotes average bits per coefficient. DG+Q12 applies 12-bit post-quantization to the decoded DG output. ``event'' denotes one complete encoding--decoding pass for a coefficient update.}
\label{tab:codec_proposed_marked}
\setlength{\tabcolsep}{1.5pt}
\begin{tabular}{@{}lccccc@{}}
\toprule
Codec & bpc & Future (\%) & Uplink (MB) & Ratio & ms/event \\
\midrule
DG FP32
& 32.00
& $48.84\!\pm\!3.86$
& $65.55$
& $1.00\times$
& 0.14 \\
DG+Q12
& 12.13
& $48.80\!\pm\!3.49$
& $24.85$
& $2.64\times$
& 11.09 \\
LRSUQ ($b=4$)
& 12.01
& $48.97\!\pm\!3.18$
& $24.59$
& $2.67\times$
& 13.97 \\
\bottomrule
\end{tabular}
\end{table}

\section{Conclusion}
\label{sec:conclusion}
In this paper, we reformulate record-private hypernetwork-based PFL around one-shot client conditioning and repeated coefficient-space adaptation. Coefficient-space adaptation preserves utility for seen and unseen clients, while radial evolution captures most of the factorization gain under privacy noise. Finally, LRSUQ further reduces the measured protected uplink by $2.67\times$ relative to FP32 DG while preserving the exact decoded Gaussian channel.

\bibliographystyle{IEEEbib}
\bibliography{refs}

\iflongversion
\clearpage
\appendix
\twocolumn[{
  \centering\normalfont
  \fontsize{11pt}{13pt}\selectfont
  \bfseries APPENDICES\par
  \vspace{15pt}
}]

\section{Fixed Coefficient Space and Restricted Gradients}
\label{app:coefficient-space}

This appendix expands the coefficient-space construction used in
Sec.~\ref{sec:private-adaptation}.  The matrix
$P\in\mathbb R^{d_\theta\times k}$ has orthonormal columns,
$P^\top P=I_k$, and therefore defines the $k$-dimensional adaptation
subspace
\begin{equation}
    \mathcal U
    =
    \operatorname{range}(P)
    =
    \{Pa:a\in\mathbb R^k\}
    \subset \mathbb R^{d_\theta}.
    \label{eq:app-subspace}
\end{equation}
Thus the columns of $P$ form an orthonormal basis of $\mathcal U$, not of
the entire $d_\theta$-dimensional model space.  Every personalized model
satisfies
\[
    \theta_i-\theta_{\rm ref}=Pa_i\in\mathcal U,
\]
so the method restricts client-specific adaptation to the affine space
$\theta_{\rm ref}+\mathcal U$.

\subsection{Public construction of the adaptation basis}

The basis is generated once from public randomness and is independent of
all client data.  One concrete realization, used in our experiments, is a
signed-partition construction.  Let
$\{\mathcal I_j\}_{j=1}^k$ be disjoint nonempty sets of model coordinates
whose union is the set of coordinates assigned to the adaptation space,
and let $s_q\in\{-1,+1\}$ be public random signs.  A column of $P$ can be
written as
\begin{equation}
    (p_j)_q
    =
    \begin{cases}
        s_q/\sqrt{|\mathcal I_j|}, & q\in\mathcal I_j,\\
        0, & q\notin\mathcal I_j .
    \end{cases}
    \label{eq:app-signed-basis}
\end{equation}
Disjoint supports imply $p_j^\top p_{j'}=0$ for $j\neq j'$, while the
normalization gives $\|p_j\|_2=1$.  Hence $P^\top P=I_k$.  Since $P$ is
fixed before any private data are observed, constructing the adaptation
space consumes no privacy budget.

The orthonormality also gives
\begin{equation}
    \|Pa\|_2^2
    =
    a^\top P^\top Pa
    =
    \|a\|_2^2,
    \label{eq:app-isometry}
\end{equation}
so coefficient norm and model-displacement norm coincide inside
$\mathcal U$.

\subsection{Coefficient gradients as exact restricted gradients}

Equation~\eqref{eq:record-gradient} is not a heuristic projection of a
gradient computed elsewhere.  It is the exact gradient of the loss with
respect to the coordinates of the affine adaptation space.  Let
\[
    L(a;z)
    =
    \ell(\theta_{\rm ref}+Pa;z).
\]
For an infinitesimal perturbation $da$,
\[
    dL
    =
    \nabla_\theta\ell(\theta_{\rm ref}+Pa;z)^\top P\,da.
\]
Therefore
\begin{equation}
    \nabla_a L(a;z)
    =
    P^\top
    \nabla_\theta
    \ell(\theta_{\rm ref}+Pa;z).
    \label{eq:app-chain-rule}
\end{equation}
Evaluating at $a=a_i$ gives
\begin{equation}
    g_i(z)
    =
    P^\top\nabla_\theta\ell(\theta_i;z).
    \label{eq:app-coeff-gradient}
\end{equation}
A coefficient-space descent step consequently induces the model-space
direction
\[
    -P g_i(z)
    =
    -PP^\top\nabla_\theta\ell(\theta_i;z),
\]
where $PP^\top$ is the orthogonal projector onto $\mathcal U$.  Thus the
coefficient update is exactly the full gradient restricted to the fixed
adaptation subspace.
This restriction does not change the privacy parameter merely by changing
dimension.  For an isotropic Gaussian vector
$\xi\sim\mathcal N(0,\sigma^2I_d)$,
\begin{equation}
    \mathbb E\|\xi\|_2^2=d\sigma^2.
    \label{eq:app-noise-energy}
\end{equation}
Hence, once sensitivity and $\sigma$ are fixed, reducing the released
dimension reduces the total expected perturbation energy and communication burden.
The $k$-sweep in Fig.~\ref{fig:triptych}(a) measures the opposing capacity
cost of making the adaptation subspace too small.

\section{Sensitivity and Privacy Accounting}
\label{app:privacy}

We use the replace-one record adjacency defined in
Sec.~\ref{sec:private-adaptation}.  The one-shot conditioning set
$\mathcal C_i$ and the round-dependent gradient minibatch
$\mathcal B_{i,t}$ play different roles.  The former is used once to form
the private context, whereas the latter is used for repeated
coefficient-gradient releases.  The accounting below composes the two
mechanisms and does not require these sets to be disjoint.

\subsection{One-shot context sensitivity}

Define
\[
    f_c(D_i)
    =
    \frac{1}{m_c}
    \sum_{(x,y)\in\mathcal C_i}\phi(x).
\]
Under replace-one adjacency, at most one summand changes from $\phi(x)$ to
$\phi(x')$.  Since $\|\phi(x)\|_2\leq1$,
\begin{align}
    \|f_c(D_i)-f_c(D_i')\|_2
    &=
    \frac{1}{m_c}
    \|\phi(x)-\phi(x')\|_2 \nonumber\\
    &\leq
    \frac{\|\phi(x)\|_2+\|\phi(x')\|_2}{m_c}
    \leq
    \frac{2}{m_c}.
    \label{eq:app-context-sensitivity}
\end{align}
Thus $\Delta_c=2/m_c$.

\subsection{Coefficient-gradient sensitivity}

For a fixed round, define
\[
    f_g(D_i)
    =
    \frac{1}{m_g}
    \sum_{z\in\mathcal B_{i,t}}
    \operatorname{clip}_{C_g}(g_i(z)).
\]
Every clipped vector has norm at most $C_g$.  Replacing one record can
therefore change at most one summand, giving
\begin{equation}
\begin{aligned}
    &\|f_g(D_i)-f_g(D_i')\|_2 \\
    &\quad\leq \frac{1}{m_g}\left(
      \|\operatorname{clip}_{C_g}(g_i(z))\|_2
      +\|\operatorname{clip}_{C_g}(g_i(z'))\|_2\right) \\
    &\quad\leq \frac{2C_g}{m_g}.
\end{aligned}
\label{eq:app-gradient-sensitivity}
\end{equation}
Hence $\Delta_g=2C_g/m_g$.

\subsection{RDP composition and conversion}

A Gaussian mechanism with $\ell_2$ sensitivity $\Delta$ and noise standard
deviation $\sigma$ satisfies order-$\alpha$ R\'enyi differential privacy
(RDP)
\begin{equation}
    \varepsilon_{\rm RDP}(\alpha)
    =
    \frac{\alpha\Delta^2}{2\sigma^2},
    \qquad \alpha>1
    \label{eq:app-gaussian-rdp}
\end{equation}
\cite{Mironov17}.  More generally, if the context and repeated gradient
mechanisms use $\sigma_c$ and $\sigma_g$, respectively, then a client
participating in $T_i$ private update rounds has
\begin{equation}
    \varepsilon_{i,\rm RDP}(\alpha)
    =
    \alpha
    \left(
        \frac{\Delta_c^2}{2\sigma_c^2}
        +
        \frac{T_i\Delta_g^2}{2\sigma_g^2}
    \right).
    \label{eq:app-rdp-general}
\end{equation}
The main text uses the common-noise setting
$\sigma_c=\sigma_g=\sigma$, for which
\[
    \varepsilon_{i,\rm RDP}(\alpha)=\alpha A_i,
    \qquad
    A_i
    =
    \frac{\Delta_c^2+T_i\Delta_g^2}{2\sigma^2}.
\]

The standard RDP conversion gives, for every $\alpha>1$,
\begin{equation}
    \varepsilon_i(\alpha,\delta)
    =
    \alpha A_i
    +
    \frac{\log(1/\delta)}{\alpha-1}.
    \label{eq:app-rdp-conversion}
\end{equation}
Let $\Lambda=\log(1/\delta)$.  Differentiating with respect to the continuous
order gives
\[
    \frac{\partial\varepsilon_i}{\partial\alpha}
    =
    A_i-\frac{\Lambda}{(\alpha-1)^2}.
\]
The minimizing order is
\begin{equation}
    \alpha_i^\star
    =
    1+\sqrt{\frac{\Lambda}{A_i}},
    \label{eq:app-optimal-order}
\end{equation}
which yields
\begin{equation}
    \varepsilon_i
    =
    A_i+2\sqrt{A_i\Lambda}.
    \label{eq:app-epsilon-closed-form}
\end{equation}
This recovers Eq.~\eqref{eq:privacy-accounting}.

For completeness, the same expression can be inverted to calibrate the
common noise level to a target $(\varepsilon,\delta)$.  Solving
$\varepsilon=A+2\sqrt{A\Lambda}$ gives
\begin{equation}
    A^\star
    =
    \left(
        \sqrt{\Lambda+\varepsilon}-\sqrt {\Lambda}
    \right)^2.
    \label{eq:app-target-A}
\end{equation}
If $T_{\max}=\max_iT_i$, the corresponding worst-client calibration is
\begin{equation}
    \sigma
    =
    \sqrt{
        \frac{\Delta_c^2+T_{\max}\Delta_g^2}
             {2A^\star}
    }.
    \label{eq:app-noise-calibration}
\end{equation}
We take the maximum over client participation counts because the schedule
is fixed independently of the data but need not give every client the same
number of private update events.  We do not invoke amplification by client
or record subsampling.

An unseen client in the deployment protocol releases only its one-shot
context and performs no private coefficient-gradient events or local
fine-tuning.  Its subsequent generated model is post-processing of that
private context and the already learned server parameters.

\section{Detailed Derivation of the Factorized Optimization Geometry}
\label{app:geometry}

This section expands the argument in Sec.~\ref{sec:geometry}.  Let
$a_i\in\mathbb R^k$, $h_i\in\mathbb R^r$,
$W\in\mathbb R^{k\times r}$, and let $\psi\in\mathbb R^{d_\psi}$ denote
the MLP parameters.  When taking derivatives, we identify the matrix $W$
with its column-stacked vectorization
$\operatorname{vec}(W)\in\mathbb R^{kr}$ and write
\[
    \omega
    =
    \big(
        \operatorname{vec}(W),\psi
    \big).
\]
The notation $\operatorname{vec}(W)$ simply stacks the columns of $W$
into one vector so that all trainable generator parameters can be treated
as a single coordinate vector.

\subsection{From the server update to the induced coefficient kernel}

The contribution of client $i$ to the server parameter step is
\[
    \Delta\omega^{(i)}
    =
    -\frac{\eta}{|\mathcal S_t|}
    J_i^\top\widetilde g_{i,t},
    \qquad
    J_i
    =
    \frac{\partial a_i}{\partial\omega}.
\]
A first-order Taylor expansion of the coefficients generated for client
$j$ gives
\begin{align}
    a_j(\omega+\Delta\omega^{(i)})
    &=
    a_j(\omega)
    +
    J_j\Delta\omega^{(i)}
    +
    O(\|\Delta\omega^{(i)}\|_2^2),
\end{align}
and therefore
\begin{equation}
    \Delta a_j^{(i)}
    \approx
    -\frac{\eta}{|\mathcal S_t|}
    J_jJ_i^\top
    \widetilde g_{i,t}.
    \label{eq:app-induced-kernel}
\end{equation}
This identifies
$K_{ji}=J_jJ_i^\top$ as the local coefficient-space coupling induced by
the shared generator.  For $j\neq i$, $K_{ji}$ is generally not symmetric,
but $K_{ij}=K_{ji}^\top$.  For $j=i$, $K_{ii}=J_iJ_i^\top$ is positive
semidefinite.

\subsection{Direct differentiation of the two parameter blocks}

Partition the Jacobian according to the two trainable parameter blocks,
\[
    J_i
    =
    \left[
        J_i^{(W)}
        \;\;
        J_i^{(\psi)}
    \right].
\]
Since $a_i=Wh_i$, the standard vectorization identity
\[
    Wh_i
    =
    (h_i^\top\otimes I_k)\operatorname{vec}(W)
\]
gives
\begin{equation}
    J_i^{(W)}
    =
    h_i^\top\otimes I_k.
    \label{eq:app-W-jacobian}
\end{equation}
Let
\[
    H_i
    =
    \frac{\partial h_i}{\partial\psi}
    \in\mathbb R^{r\times d_\psi}.
\]
The chain rule gives
\begin{equation}
    J_i^{(\psi)}
    =
    WH_i.
    \label{eq:app-psi-jacobian}
\end{equation}
Hence
\begin{align}
    K_{ji}
    &=
    J_j^{(W)}(J_i^{(W)})^\top
    +
    J_j^{(\psi)}(J_i^{(\psi)})^\top
    \nonumber\\
    &=
    (h_j^\top\otimes I_k)(h_i\otimes I_k)
    +
    WH_jH_i^\top W^\top
    \nonumber\\
    &=
    \langle h_j,h_i\rangle I_k
    +
    WH_jH_i^\top W^\top.
    \label{eq:app-kernel-derivation}
\end{align}
Equation \eqref{eq:app-kernel-derivation} is obtained directly by partitioning the
Jacobian by parameter block and multiplying the blocks.

The first term is isotropic in coefficient space.  For $i=j$, it is the
positive scalar
$\|h_i\|_2^2$ times the identity.  The second term has rank at most $r$
and can be direction dependent.  In the self-coupling case,
\begin{equation}
    K_{ii}
    =
    \|h_i\|_2^2I_k
    +
    WH_iH_i^\top W^\top,
    \label{eq:app-self-kernel}
\end{equation}
where the second term is also positive semidefinite.  Thus the
factorization supplies a positive isotropic component together with an
additional low-rank, potentially anisotropic correction.

\subsection{Radial evolution and the ablation controls}

Write the hidden representation as
\begin{equation}
    h_i
    =
    \rho_i v_i,
    \qquad
    \rho_i=\|h_i\|_2,
    \qquad
    \|v_i\|_2=1.
    \label{eq:app-radial}
\end{equation}
We use \emph{radial evolution} to mean that the hidden direction remains
fixed, $v_i=v_i^0$, while only the magnitude $\rho_i$ changes.  Under this
restriction, the isotropic part of the self-kernel becomes
\begin{equation}
    \|h_i\|_2^2I_k
    =
    \rho_i^2I_k.
    \label{eq:app-radial-gain}
\end{equation}
Thus radial motion in hidden space induces an adaptive scalar gain on all
coefficient directions without rotating the hidden representation.

This interpretation explains the controls in
Sec.~\ref{sec:geometry-ablation}.  \emph{Radial} suppresses hidden-direction
evolution while retaining the factorized parameterization, testing whether
magnitude adaptation is sufficient for most of the gain.  \emph{Replay}
removes the factorization and applies only the recorded scalar sequence
$\gamma_t=\|h_t\|_2^2$ to Direct.  Replay therefore tests a stronger
hypothesis: whether the factorized benefit can be reduced to a prescribed
scalar step-size schedule in the unfactorized parameterization.  The two
controls are not equivalent because Radial retains the coupled factorized
dynamics, whereas Replay does not.

Finally, Eq.~\eqref{eq:app-induced-kernel} is a local first-order
description.  We use it to interpret how a noisy coefficient update is
transferred through the shared generator.  It is not a global convergence
theorem.

\section{LRSUQ, Exact Gaussian Channel Simulation, and Expected Code Length}
\label{app:lrsuq}
The communication construction in Sec.~\ref{sec:lrsuq} should be interpreted
as exact simulation of the decoded Gaussian channel using variable-length
messages with finite expected code length. Specifically, if $L$ denotes the
number of transmitted bits for one encoded message, the relevant guarantee is
$\mathbb E[L]<\infty$. This does not imply a uniform finite-length bound on
every realization.

\subsection{From subtractive dithering to prescribed error laws}

Classical subtractive dithered quantization uses shared randomness to make
quantization error independent of the source.  The corresponding
input-independence principle is commonly expressed through the Crypto Lemma
\cite{zamir2014lattice}.  In conventional lattice dithering, however, the
error law is tied to the quantization cell.

LRSUQ uses rejection sampling on top of universal quantization to shape the
input-independent quantization error to a prescribed continuous target law
\cite{ling2025lrsuq}.  In our application the target for each $b$-dimensional
block is
\[
    \mathcal N(0,\sigma^2I_b).
\]
With independent shared randomness across blocks, concatenating the decoded
blocks gives
\begin{equation}
    \widehat u
    =
    u+e,
    \qquad
    e\sim\mathcal N(0,\sigma^2I_{d_u}),
    \qquad
    e\perp u.
    \label{eq:app-lrsuq-gaussian}
\end{equation}
The exact prescribed error law in
Eq.~\eqref{eq:app-lrsuq-gaussian} is the LRSUQ property
\cite{ling2025lrsuq}.

\subsection{Equality of the decoded channel and the DP consequence}

Let
\[
    M_{\rm DG}(u)=u+\xi,
    \qquad
    \xi\sim\mathcal N(0,\sigma^2I_{d_u}),
\]
and let $M_{\rm LRSUQ}(u)$ denote the decoded LRSUQ output.  From
Eq.~\eqref{eq:app-lrsuq-gaussian},
\begin{equation}
    M_{\rm LRSUQ}(u)
    \overset{d}{=}
    M_{\rm DG}(u)
    \qquad
    \text{for every fixed }u,
    \label{eq:app-channel-equality}
\end{equation}
where $\overset{d}{=}$ denotes equality in distribution.  This is an
equality of conditional output laws, not an assertion that the two
mechanisms generate identical sample paths.

For any measurable decoded-output set $\mathcal A$,
\[
    \Pr[M_{\rm LRSUQ}(u)\in\mathcal A]
    =
    \Pr[M_{\rm DG}(u)\in\mathcal A].
\]
Therefore any $(\varepsilon,\delta)$ inequality established for the decoded
DG mechanism also holds for decoded LRSUQ with the same sensitivity and
noise scale.  This is why replacing DG with LRSUQ does not require a new
decoded-output privacy accountant.

The claim is deliberately restricted to the trusted decoded output.  We do
not claim local DP for the raw variable-length codeword, its length, or the
shared randomness.  These quantities remain inside the trusted
encoder--decoder boundary in our central-DP threat model.

\subsection{Relation to post-quantization baselines}

A codec such as DG+Q12 first generates an already private Gaussian output and
then quantizes that output.  The additional quantization is DP-preserving
post-processing, but it changes the decoded channel by adding a second
distortion layer.  LRSUQ instead uses its quantization error to realize the
required Gaussian perturbation itself. The matched-rate experiment therefore
compares two different communication constructions at similar realized rates:
post-quantization of a Gaussian-private release versus direct variable-length
simulation of the Gaussian-private channel.

\fi

\end{document}